\documentclass[letterpaper, 10 pt, conference]{ieeeconf}  % Comment this line out
\IEEEoverridecommandlockouts                              % This command is only
\usepackage{graphicx}
\usepackage{tabularx}
\usepackage{comment}
\usepackage[utf8]{inputenc}
\usepackage[T1]{fontenc}
\usepackage{subcaption}

\newsavebox{\twosubbox}
\usepackage{xcolor, hyperref}
\usepackage{color,soul}
\usepackage{float}
\usepackage{graphics} % for pdf, bitmapped graphics files
\usepackage{amsmath} % assumes amsmath package installed

\title{\LARGE \bf
Evaluating Pointing Gestures for Target Selection in Human-Robot Collaboration
}

\author{Noora Sassali$^{1}$ and Roel Pieters$^{1}$% <-this % stops a space
\thanks{$^{1}$Cognitive Robotics group, Unit of Automation Technology and Mechanical Engineering, Tampere University, 33720, Tampere, Finland; {\tt\small firstname.surname@tuni.fi}}%
}

\begin{document}

\maketitle
\thispagestyle{empty}
\pagestyle{empty}

%%%%%%%%%%%%%%%%%%%%%%%%%%%%%%%%%%%%%%%%%%%%%%%%%%%%%%%%%%%%%%%%%%%%%%%%%%%%%%%%
\begin{abstract}

Pointing gestures are a common interaction method used in Human-Robot Collaboration for various tasks, ranging from selecting targets to guiding industrial processes. This study introduces a method for localizing pointed targets within a planar workspace. The approach employs pose estimation, and a simple geometric model based on shoulder-wrist extension to extract gesturing data from an RGB-D stream. The study proposes a rigorous methodology and comprehensive analysis for evaluating pointing gestures and target selection in typical robotic tasks. In addition to evaluating tool accuracy, the tool is integrated into a proof-of-concept robotic system, which includes object detection, speech transcription, and speech synthesis to demonstrate the integration of multiple modalities in a collaborative application. Finally, a discussion over tool limitations and performance is provided to understand its role in multimodal robotic systems.
All developments are available at: \textit{\footnotesize \url{https://github.com/NMKsas/gesture_pointer.git}}
\end{abstract}

%%%%%%%%%%%%%%%%%%%%%%%%%%%%%%%%%%%%%%%%%%%%%%%%%%%%%%%%%%%%%%%%%%%%%%%%%%%%%%%%
\section{Introduction}\label{sec:intro}

Deictic gestures are a natural way to interact with the world to identify objects of interest \cite{wachs_vision-based_2011}. In collaborative robotic systems, pointing gestures can be used as a powerful tool to perform decision-making, such as target selection. Using gestures has its advantage especially in industrial environments, where interpreting speech commands can be difficult due to noise. 

\begin{figure}[!t]
\centering
\includegraphics[width=6.0cm]{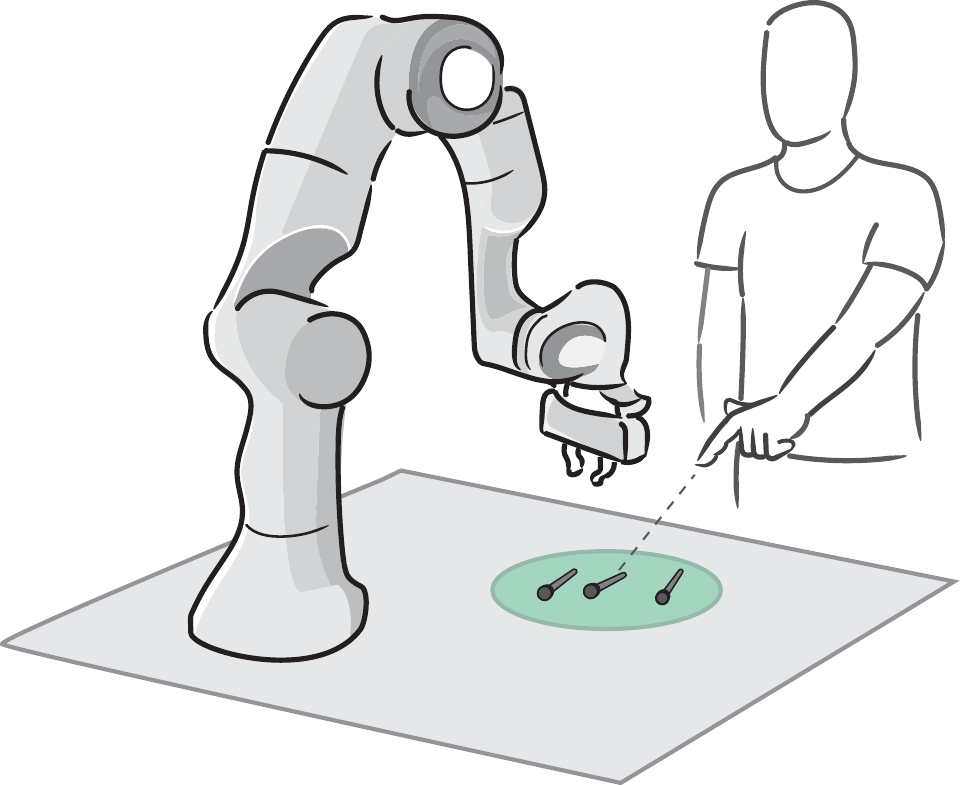}
\caption{Selecting a target with pointing gestures.}
\label{fig:gesturing_concept}
\end{figure}

Localizing pointing gestures creates a new layer of information that can be used as a basis for gestural algorithms in collaborative systems. This study presents a vision-based approach that utilizes RGB-D stream instead of wearable sensors, and fulfills the real-time performance requirements of the industry cost-efficiently. The tool neither requires excessive training nor physical strain from the users, offering an ergonomic alternative for traditional user interfaces. An illustration of the concept is shown in Fig. \ref{fig:gesturing_concept}. 

While pointing gestures have been widely used in robotics and the methodology follows its predecessors \cite{hu_augmented_2022, jevtic_personalized_2019, medeiros_human-drone_2020}, the focus of this work is to form a basis for gestural communication and its integration to a larger system. The study provides a structured evaluation for pointing accuracy within a planar workspace and features two simple algorithms to implement target selection in pick and place tasks. A proof-of-concept integration into a robotic system, which includes object detection, speech interpretation and feedback modules, is presented. Finally, an analysis over tool's performance is provided, discussing tool's shortcomings and relevancy in real-world applications.  

The paper is structured as follows. Section \ref{sec:related_work} provides a brief background and reviews the related works. Section \ref{sec:system} gives an overview of the developed system and relevant theory behind it. Section \ref{sec:evaluation} describes how the system was tested. Section \ref{sec:results_discussion} delivers the results and discusses the limitations of the tool. Finally, the paper concludes in section \ref{sec:conclusion}.  

\section{Related Work}\label{sec:related_work}

% Brief history 
Pointing gestures are traditionally localized by recognizing and locating the human in the scene, and tracking meaningful parts of the body such as head and hands to determine the pointing direction. Early gestural implementations for target localization relied on classical image processing techniques to extract and process feature maps from the scene. \cite{kahn_gesture_1996} 
Expansion to geometric framework through the application of stereo vision allowed recovering the line along which the index finger is pointing and localizing the intersection point on a two-dimensional plane \cite{cipolla_human-robot_1996}. The sequential and temporal nature of gesturing has been investigated using different state models, such as Hidden Markov Models \cite{nickel_visual_2007, droeschel_learning_2011}, to recognize the exact occurrence the gesture.

% Recent advances / SOTA 
Currently pointing gestures can be efficiently exploited with hand-held sensors, such as wristbands with Intertial Measurement Units (IMU) \cite{gromov_proximity_2019, abbate_pointit_2022, kang_minuet_2019}, but many works prefer computer vision due to its low cost and non-intrusive nature.
As the mainstream of vision-based works estimate pointing gestures by projecting a vector between localized body parts, the development tracks closely the advances in pose estimation. Commonly used pairs for determining the pointing direction include e.g., head--finger \cite{li_visual_2010}, shoulder--finger \cite{lorentz_pointing_2023}, and elbow--wrist \cite{hu_augmented_2022, medeiros_human-drone_2020} combinations. The limitations of defining pointed targets via linear extrapolation are discussed by Herbort and Kunde in \cite{herbort_spatial_2016}. Alternative approaches without pose estimation include unsupervised learning \cite{jirak_solving_2021}, and probabilistic appearance-based frameworks \cite{shukla_probabilistic_2015}.

% Overview in HRC
Applications of pointing gestures are numerous in Human-Robot Collaboration (HRC). Pointing gestures have been used for directing mobile robots to specified locations \cite{michal_foundations_2017, moh_gesture_2019}. Similarly, the modality has been utilized in aerial robotics to guide drones towards their intended targets \cite{gromov_proximity_2019, medeiros_human-drone_2020}. Other applications include requesting footwear in robot-assisted dressing \cite{jevtic_personalized_2019} and assigning pick and place tasks to the robot \cite{lorentz_pointing_2023}. Industrial examples feature pointing gestures as a mean to select objects on conveyor belts \cite{abbate_selecting_2022} and to localize targets in construction site tasks \cite{yoon_laserdex_2024}. Pointing gestures are often accompanied by speech in multimodal systems and combining modalities facilitates the interaction between the human and the robot \cite{kang_minuet_2019, constantin_multimodal_2023}.

% Motivation 
In this paper, pose estimation with shoulder-wrist pair extension is used for localizing the gestured target within a planar workspace. The geometric, vision-based approach is a low-cost and simple alternative for wearable devices and complex models. Extending previous contributions, the study proposes a rigorous methodology and comprehensive analysis for evaluating directed gestures in common robotic tasks. Object detection, speech detection and synthesis are included in the final proof-of-concept robotic system, to demonstrate the integration of modalities in a collaborative application. 

\section{System} \label{sec:system}

\begin{figure}[!t]
\centering
\includegraphics[width=8.0cm]{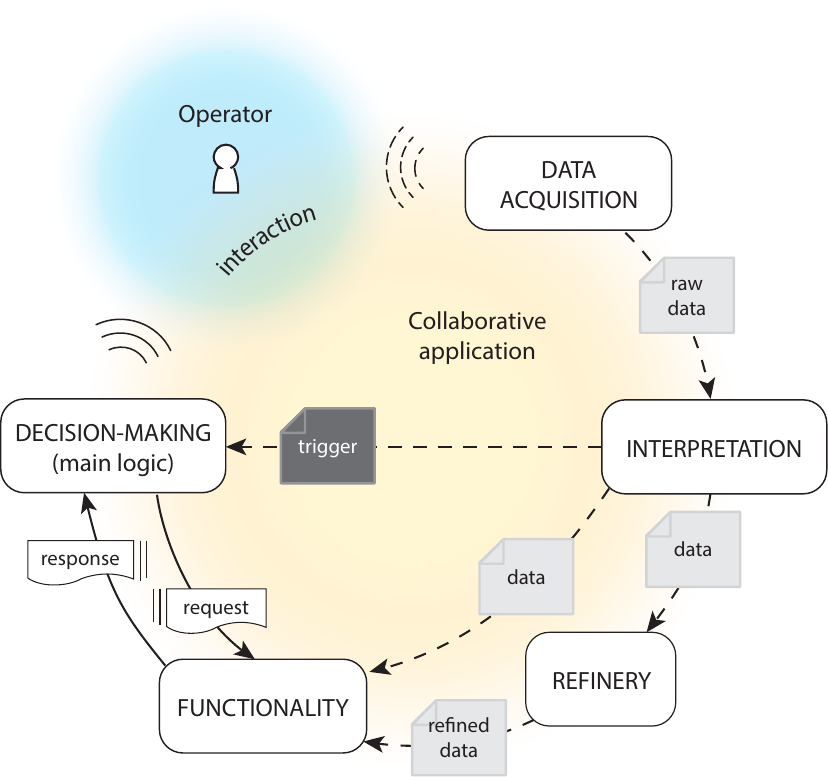}
\caption{Modeled collaborative application.}
\label{fig:modules}
\end{figure}

A collaborative application can be modeled as a distributed robotic system using modules that fall into five distinct categories: data acquisition, interpretation, refinery, decision-making and functionality. The upper level architecture is illustrated in Fig. \ref{fig:modules}. Data acquisition provides the means to sense user interactions and delivers raw data for further interpretation. Interpretation is responsible for relaying the data and sending trigger events to the main logic. Refinery modules post-process the interpreted data to create new value for the system. Both interpreted and refined data are used by functionality modules, which provide functions to the main logic using client-server communication. The main logic uses client interfaces to give operator feedback and mediate the requested actions to their corresponding functional units. 

The developed gesturing tool falls into the category of refinery, as it uses sensed data to produce new value for the application. As the tool itself does not determine how the localized pointing gesture is used, a corresponding functional unit to perform target selection was created. Due to compatibility requirements with  Franka Emika Panda robot in integration tests, ROS1 Noetic was selected for the node development.

\subsection{Pointing gestures} \label{subsec:pointing_gestures}

A rectangular workspace can be mathematically modeled as a 3D plane, using three non-collinear points in 3D space. The plane equation is
\begin{align}
    ax + by + cz + d = 0 \label{eq:general_plane},
\end{align}

where $(a,b,c)$ are the values of plane normal $\mathbf{n}$, derived from the cross product of two vectors along the plane $\Vec{P_1 P_2}$ and $\Vec{P_2 P_3}$. The constant term $d$ can be defined by taking a dot product between the normal and the third known point $P_3$,
\begin{align}
    d &= -\langle \mathbf{n}\; , P_3\; \rangle. \label{eq:plane_inner_product}
\end{align}

A gesturing workspace requires 3D corner coordinates $P_1$, $P_2$ and $P_3$ for the desired workplane. The fourth point $P_4$ is collected to complete plane limits. Corners can be visually selected from a 2D image and then deprojected into 3D coordinates using depth stream and intrinsic camera parameters. Alternatively, the user can define plane corners with ArUco markers of pre-defined size using RGB stream. Both of the methods are implemented for the tool.

The gestured point in the plane can be approximated by extending the line between shoulder and wrist points $P_s$,$P_w$; assuming the 3D points of the joints are known. The pointed target is an intersection point between the line and the defined plane. The intersection point $P_i$ can be defined, 
\begin{align}
    P_i &= P_s + t\cdot\Vec{P_sP_w},  \label{eq:intersection}
\end{align}

where $t$ is the scaling factor to move from shoulder point $P_s$ along the directional vector $\Vec{P_sP_w}$, until plane intersection is reached. The factor $t$ can be solved,
\begin{align}
    t &= \frac{-(\langle \mathbf{n}\; , P_s\; \rangle + d)}{\langle \mathbf{n}\; ,\Vec{P_sP_w}\; \rangle}\label{eq:factor_t_dot_product}.    
\end{align}

\begin{figure} [b]
\centering
\includegraphics[width=8cm]{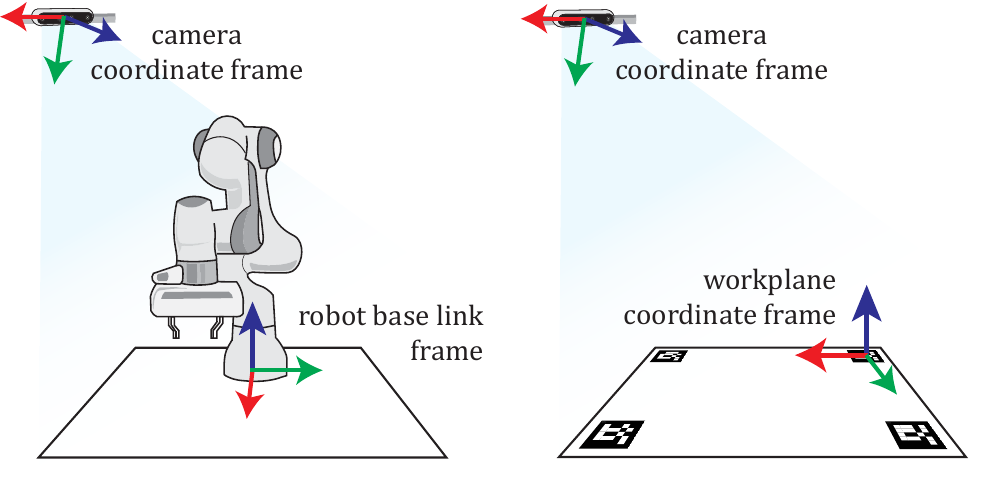}
\caption{Creating a separate reference frame for gesturing tool helps interpreting the pointed location with respect to the used workstation.}
\label{fig:coordinate_frames}
\end{figure}

\begin{figure*}[ht]
\centering
\subcaptionbox{The quantitative test setup\label{fig:test_setup_quantitative}}{%
  \includegraphics[height=0.22\linewidth]{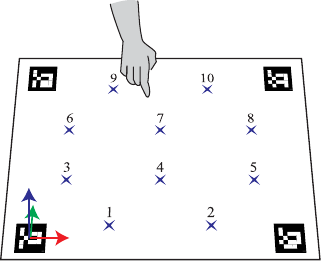}
}\hspace{0.5cm}
\subcaptionbox{The pick test setup \label{fig:test_setup_pick}}{%
  \includegraphics[height=0.22\linewidth]{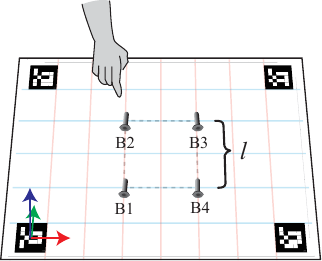} 
}\hspace{0.5cm}
\subcaptionbox{The place test setup \label{fig:test_setup_place}}{%
  \includegraphics[height=0.22\linewidth]{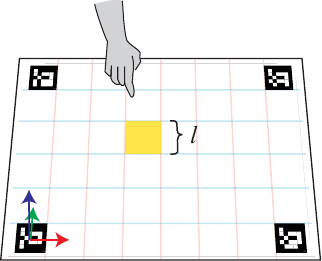} 
}\hspace{0.5cm}
\caption{
Right-handed pointing for quantitative and qualitative test setups.
\label{fig:test_setups}
}
\end{figure*}

Gesturing tool uses RGB-D stream and 2D pose estimation node implemented within OpenDR project \cite{opendr2022} to solve and publish the intersection point as ROS topic. The pose estimation is based on OpenPose \cite{openpose_2021} and detects human body keypoints as 2D image coordinates. The 2D image coordinates are de-projected into 3D points using depth stream and camera intrinsics provided by the camera. As pose estimation always introduces some noise to shoulder and wrist point estimates, the gestured position is stabilized by buffering. The published position is a running average of five samples located within the workspace.

\subsection{Coordinate frames} \label{subsec:coordinate_frames}

A robotic system can include multiple coordinate reference frames, such as joint and sensor frames. A common, fixed world coordinate frame combines the available data from different sources. In this particular use case, the robot base link forms the root node for the world coordinate system and is used as a reference frame for e.g., object detection. The camera is added to the transformation tree by performing an extrinsic camera calibration. In ROS based systems, transformations are usually handled using \texttt{tf2\_ros} library \cite{tf2_repository} and \texttt{MoveIt!} \cite{moveit_repository} plugin for calibration.

When robot base link is used as a reference frame, the gesturing plane aligns conveniently with $xy$-axis, making $z$ value constant for all points residing on the plane. With dimensionality reduction to 2D, interpreting the gestured point is intuitive with respect to the workspace. A separate coordinate frame at a workplane corner reduces the complexity similarly, and enables 2D interpretation for the plane coordinates when no robotic system is involved. Fig. \ref{fig:coordinate_frames} illustrates the analogy between the reference frame setups. 

The orientation of the workplace coordinate frame can be solved using quaternion algebra, the plane normal $\mathbf{n}$ and a directional vector from the desired corner to another.

\subsection{Snap to target module} \label{subsec:snap_to_target}

As pointing gestures are mostly used for selecting targets such as objects or locations, \texttt{snap\_to\_target} module was developed. The package forms a ROS action server interface to respond target selection client calls, falling into category of functional modules in the collaborative application (Fig. \ref{fig:modules}). The module consists of main class \texttt{SnapActionNode} to implement the action server, and an abstract class \texttt{SnapStrategy}. Each snap node has an instance of \texttt{SnapStrategy} that defines an algorithm for locking into workspace target.

The system may have multiple snap action servers running with different selection algorithms. For example, a gesture-object strategy uses pointing gesture and detection coordinates derived by an object detection model to select a target. Speech related strategies rely on direct speech input; when the operator utters the name of an object or location, a target group or an individual target is chosen. The abstraction allows creating more complex algorithms with multiple modalities as long as the common interface, abstract methods of \texttt{SnapStrategy}, are implemented. The decision-making unit can request target selection from the servers in parallel and choose the target from responses based on speed or other criteria. 

To test the gesturing tool, two simple strategies were implemented. The first strategy computes Euclidean norm between known targets and the positions published by the gesturing tool. The selection is made using mean over $N=15$ gestured point samples, given the samples' radial distance from the mean is less than $5$cm. Thresholding with radial distance ensures the gesture is stable enough to be interpreted as selection. The strategy can be used for picking up targets from the workspace.

The second strategy is used for an area selection. Similarly to previous method, the mean gestured point over $N=15$ samples is computed and used only if the radial distance from the mean remains less than $5$cm. If the mean position falls into limits of predefined areas, the selection is made. If gesturing is out of bounds, the pointed mean position is compared to existing area center points, and the nearest area is selected. The strategy can be used for selecting a place location from the workspace.

\subsection{Experimental setup} \label{subsec:experimental_setup}

\textbf{Intel RealSense D415} camera was used for implementing the gesturing tool, after performing a set of camera calibrations (dynamic target, on-chip, tare and focal length) as instructed by the manufacturer. Testing was performed using PC with Nvidia GTX 1080Ti 11GB GPU to run gesturing, snap to target, pose estimation and object detection nodes. The final integration setup included \textbf{Franka Emika Panda} 7-DOF robot with \textbf{Intel RealSense D435} as an end-effector camera. A second PC with no special hardware was used to control the robot and run some of the ROS1 nodes included in the application. 

The object detection was based on the model developed in a thesis project by Ojanen \cite{ojanen2024}. \texttt{MoveIt!} controller interface used in the system was provided by Parikka \cite{ossi_repo}. A speech transcription node developed in OpenDR project \cite{opendr2022} and \texttt{audio\_common} ROS package \cite{audio_common_repository} were used for creating the speech interpretation and feedback for the system. The tool, documentation on its use, and designed 60x80cm testing boards are provided in an open-source repository \url{https://github.com/NMKsas/gesture_pointer.git}.

\section{Evaluation} \label{sec:evaluation}

Both quantitative and qualitative testing were performed to the gesturing tool. Numerical results form the baseline accuracy for the tool, and give insight into tool performance. Qualitative testing takes the gesturing tool into context of collaborative applications by using 
\texttt{snap\_to\_target} module to study gestures in target selection. Final tests integrate the gesturing and selection methods into a real robotic application. The integration showcases how gesturing can be used as one of the modalities to perform decision-making for typical robotic actions: picking, placing and giving objects. 

It is noteworthy that all the tests were concluded by a right-handed person, which is likely to contribute to asymmetry of the results. While the gesturing tool includes a projection stream to give real-time feedback of the pointed target, it was not used for visual cues during the testing. Only shoulder-wrist combination for the tool was used, as it performed significantly better than the elbow-wrist variant in the chosen use scenario. 

\subsection{Quantitative testing} \label{subsec:quantitative_tests}

Quantitative testing measures the accuracy and precision of the tool. Ten ArUco targets were created and placed on the workspace table, as illustrated in Fig. \ref{fig:test_setup_quantitative}. Ground truths were computed as mean position over $N=100$ samples, using \texttt{aruco\_ros} \cite{aruco_ros_repository} package. The error $e$ was calculated as an Euclidean norm between the ground truth and the gestured position,
\begin{align}
    e &= \sqrt{\Delta x^2 + \Delta y^2 + \Delta z^2}\label{eq:euclidean_norm}.
\end{align}

%The error was computed both with respect to the camera frame as 3D coordinates, and with respect to the established workplane frame where $z\approx0$. 
The error was computed with respect to the camera frame as 3D coordinates. Two different methods for defining the workplane were tested separately: RGB marking the ArUco approach, RGB-D the graphical user interface based approach. After testing the framework in camera coordinate system, the tests were repeated using absolute positions with respect to the workplane frame as ground truth. The purpose of the second setup was to validate the tool setup that publishes gestured points in more intuitive workplane frame coordinates ($z=0$). 

\subsection{Qualitative testing} \label{subsec:qualitative_tests}

A set of qualitative tests were conducted using the pick and place selection methods implemented in \texttt{snap\_to\_target} module.

The pick test flow included four bolts $\{\text{B1},\text{B2},\text{B3},\text{B4}\}$ laid out on the middle of table in a square formation, with side distance $l$. The setup is illustrated in Fig. \ref{fig:test_setup_pick}. The bolt locations were defined as absolute coordinates with respect to the workplane coordinate system. The gesturing for target selection was then performed 10 times per each bolt, over a set of different distances $l=\{40, 30, 20,10,8,6,4,2\}$cm. Success rates (\%) for correctly evaluated selections were computed. 

The place test flow, shown in Fig. \ref{fig:test_setup_place}, focused on area selection. Three square targets of size $l=\{20,10,5\}$cm were moved on a workplane of size $60\times80$cm and gesturing was performed 10 times to each area. Success rate-\% and offsets from ground truth were computed. 

\subsection{Integration tests} \label{subsec:integration_tests}

Both gesturing tool and \texttt{snap\_to\_target} module were integrated into a collaborative application with 6-DoF Franka Emika Panda robotic arm. The application consists of a behavior tree-based decision-making logic and ROS servers for functionalities such as speech interpretation, speech synthesis, and robotic control. The operator could initiate \textit{pick}, \textit{place} and \textit{give} flows by uttering the corresponding command to the system. Three integration scenarios were tested separately with left and right hand, using RGB-D defined workplane. Unlike qualitative tests, integration tests included a certain degree of freedom, as none of test setups were strictly defined with respect to the workplane or camera coordinate frame. 

\begin{figure*}[ht]
\centering
\subcaptionbox{The right-handed setup \label{fig:right_handed_integration_t12}}{
  \includegraphics[height=0.22\linewidth]{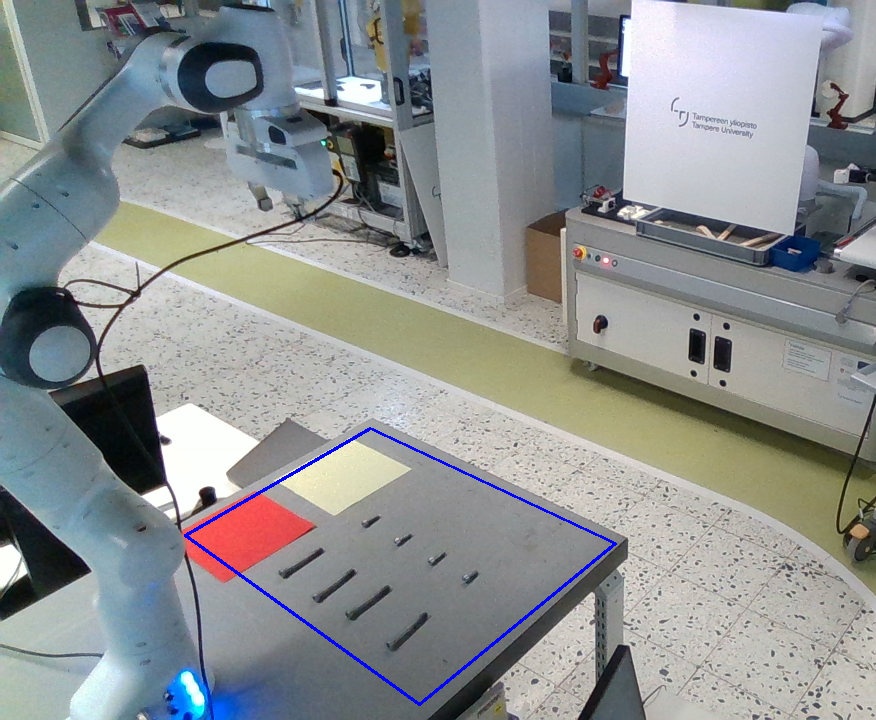} 
}\hspace{0.4cm}
\subcaptionbox{The left-handed setup
\label{fig:left_handed_integration_t12}}{
  \includegraphics[height=0.22\linewidth]{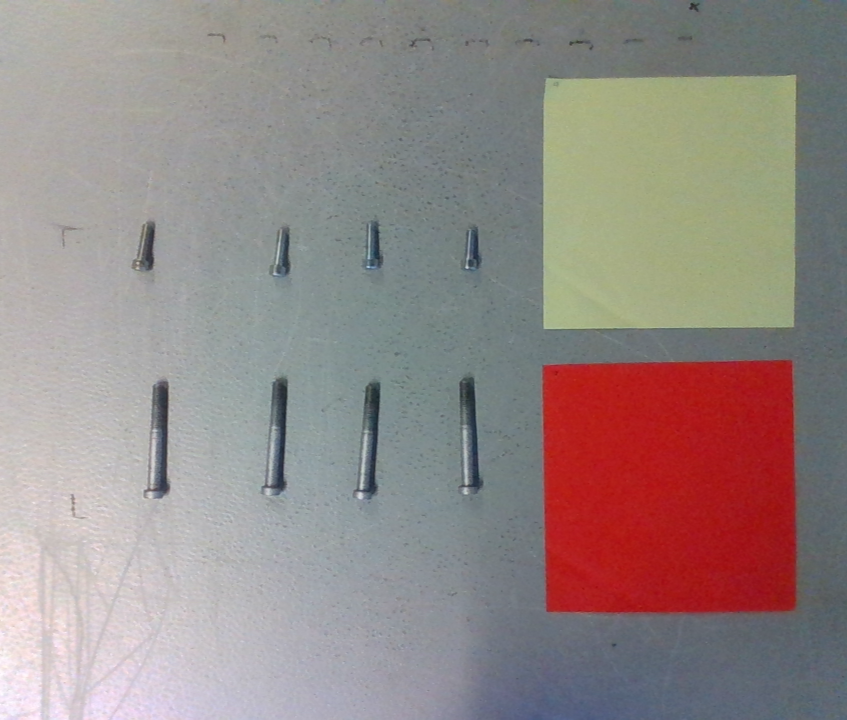}
}\hspace{0.4cm}
\subcaptionbox{The cluttered setup \label{fig:cluttered_scene}}{
  \includegraphics[height=0.22\linewidth]{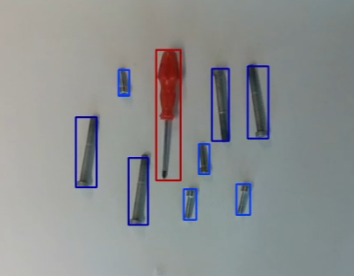} 
}\hspace{0.4cm}
\caption{
Integration test setups for task flows 1--3.
\label{fig:integration_setups}
}
\end{figure*}

In the first task flow (T1), 4 big bolts and 4 small bolts were placed on the workplane approximately 10cm apart, with two $20 \times20$cm place areas. Place areas were located on the same side as the used hand, to keep the interaction sequence natural. The operator's task was to sort the targets to their corresponding place areas, four times per each side. The setup showcases target selection by combining the gesturing tool and object detection model. Fig. \ref{fig:right_handed_integration_t12} and \ref{fig:left_handed_integration_t12} show the left- and right-handed sorting setups. 

The second task flow (T2) incorporates speech modality to select an object group before an individual target. The operator sorts the bolts just as before, but utters the group name ("big bolt", "small bolt") to first limit the set of selectable objects. After snapping to group, the individual target is selected with gesturing, following the simple Euclidean distance based snap strategy. In task flow 2, the aim is to demonstrate how combining modalities improves the decision-making.

The final integration test (T3) features 4 big bolts, 4 small bolts and a screwdriver, placed randomly on the table near each other, as seen in Fig. \ref{fig:cluttered_scene}. The final test setup takes system to its limits by increasing the difficulty of the task. The user prompts the system to give each item from the table, one by one. After each item is received, the operator returns the items back on the table in random order. The give-flow is repeated four times with both hands.

\section{Results and Discussion} \label{sec:results_discussion}

\begin{figure*} [!t]
\centering
    \includegraphics[width=7in]{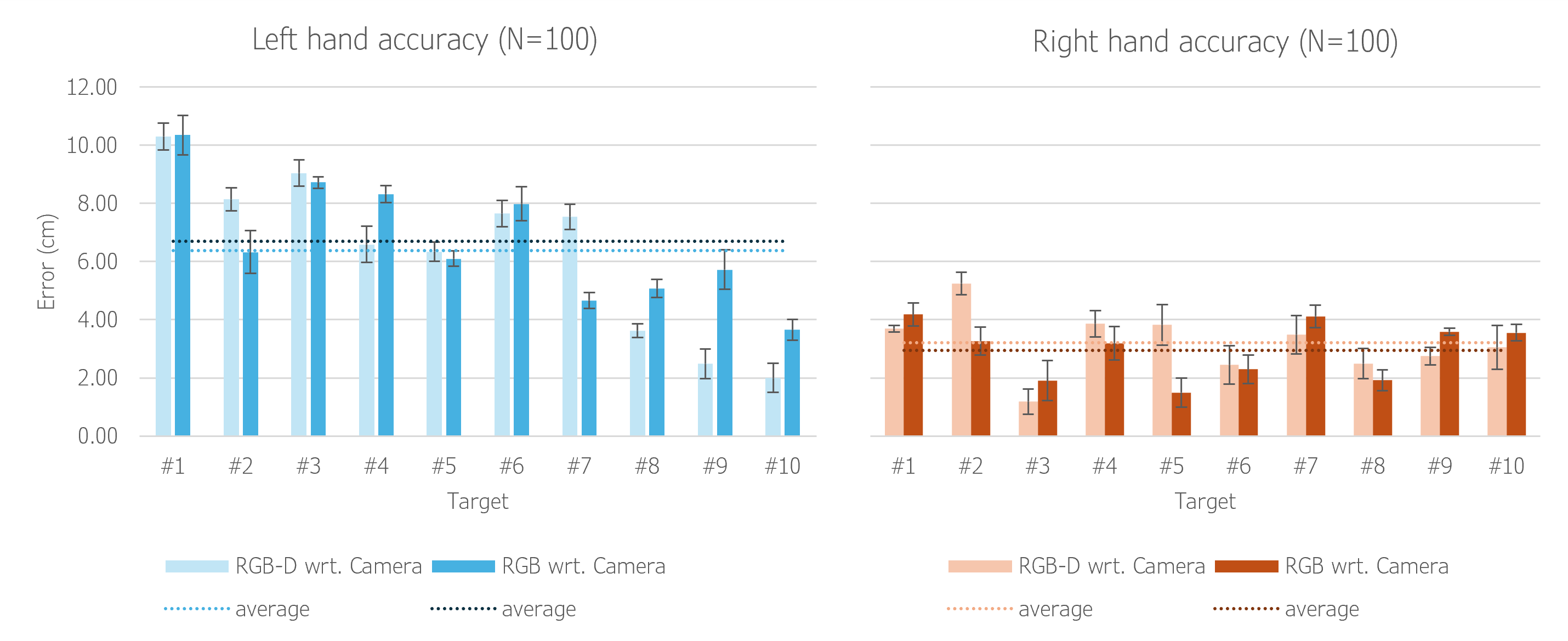}
    \caption{Numerical results for gesturing tool accuracy.}
    \label{fig:accuracy_results}
\end{figure*}

\subsection{Quantitative results}

Left and right hand results are shown in Fig. \ref{fig:accuracy_results}. For both hands, the standard deviation of error is around half a centimeter, regardless of approach. This likely results from buffering implemented within the tool. The approach-wise difference in mean errors is less than 0.4cm and can be considered insignificant. As deviations between the approaches are very subtle, it can be said the setups are equally suitable for establishing the workplane.

The dominant (right) hand is significantly more accurate in pointing tasks. Right hand reaches an average accuracy of 3.0–3.3cm, while left hand  results in the double, 6.4–6.7cm. The asymmetry is likely connected to the handedness of the user. When 3D coordinates were transformed to workplane coordinate frame ($z=0$) and compared to their absolute coordinates, the average error reduced by approximately 0.5cm for both hands. This validated using workplane coordinate frame in qualitative tests. 

The results reinforce the view that pointing gestures highly depend on pointers individual characteristics in perceptual features, as established in previous studies \cite{lorentz_pointing_2023, herbort_spatial_2016}. The performance is also dependent on the target location. Targets closer to the user are less prone to errors, while the farthest targets on the workplane tend to cause overshoot even further. 

\subsection{Pick test flow}

\begin{figure}[!b]
    \centering
    \includegraphics[width=7cm]{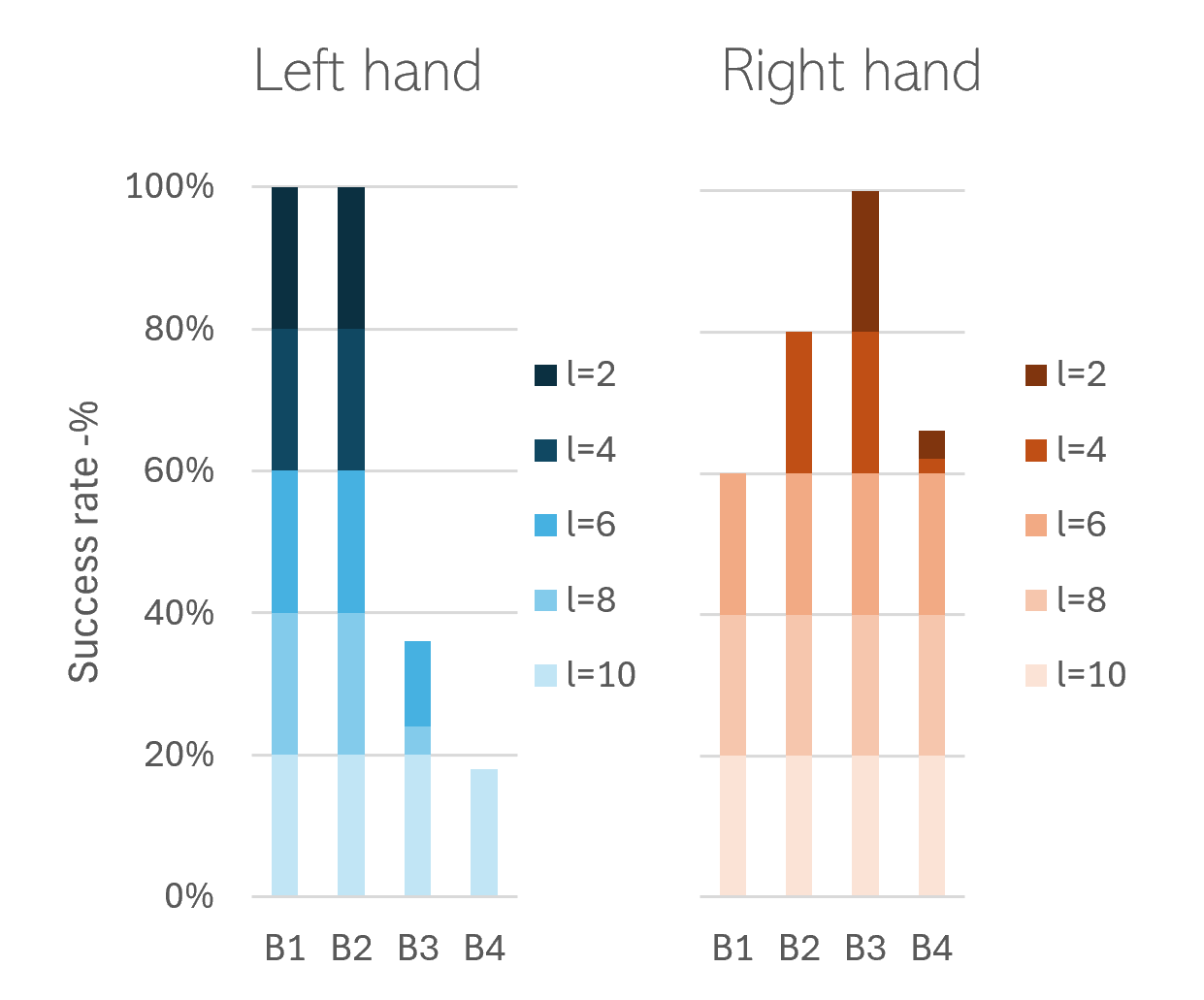}
    \caption{Pick flow - selection success rate-\% ($N=50$).}
    \label{fig:pick_flow_success}
\end{figure}

The success rate for each bolt over $N=50$ samples is illustrated in Fig. \ref{fig:pick_flow_success}. Distances $l=\{20,30,40\}$ are left out, as they resulted in 100\% success for each bolt. 

According to the numerical analysis, the average accuracy for left hand gesturing is 6.4--6.7cm. It can be hypothesized that the objects can be told apart only if $l$ exceeds 12cm between the bolts. The results indicate the opposite for bolts B1 and B2, as seen in Fig. \ref{fig:pick_flow_success}. During the testing it became evident that the test setup is susceptible for false positives, due to gesturing overshoot. As only four objects are selectable and the gesturing overshoots in favor of the objects opposite to the handedness, snapping to target succeeds even with $l=2$cm distance between the bolts.

\begin{figure*}[t]
    \centering
    \includegraphics[width=6.8in]{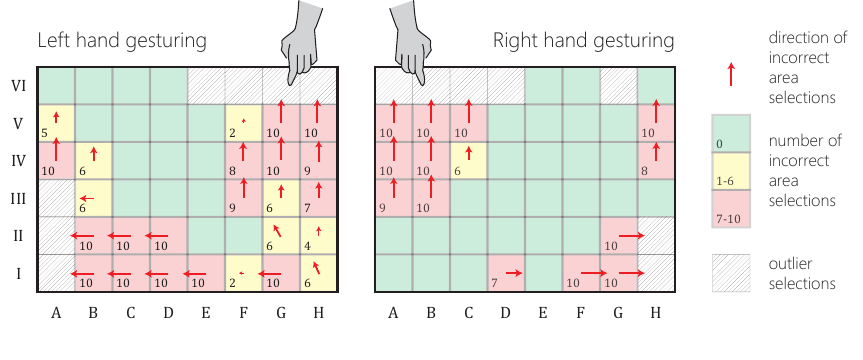}
    \caption{Place flow; area selection success $N=10$, $l=10$cm.}
    \label{fig:place_flow_success}
\end{figure*}

Considering the overshoot, bolts B3 and B4 assess the left hand performance more realistic than B1 and B2. Over $N=50$ samples, the left hand succeeds in selection almost fully for $l=10$cm, but as the distance between the bolts decreases the total success rate reaches only 36\% for B3, and 18\% for B4. The result aligns with the numerical analysis. 

Similarly, the right hand performance is best evaluated by bolts B1 and B2. The average accuracy for right-hand gesturing is 3.0–-3.3 cm, as demonstrated by numerical analysis. This suggests that target selection should be successful when the distance $l$ exceeds 6 cm. The pick flow results for B1 and B2 support the assumption, as 10 selections are made successfully over each distance  $l=\{6,8,10\}$ cm. The success rate drops for the smaller distances, and total success rate over all the samples $N=50$ is 60\% for B1 and 80\% for B2. 

\subsection{Place test flow} \label{subsec:place_test_flow_results}

The first area size, $l=20$cm, resulted in 100\% success rate for both hands. As expected, the second area size $l=10$cm already posed a challenge for the system, as successful pointing at the area center would require at least 5cm accuracy. The success rate ($l=10$cm) for the left hand was 56.67\% and for the right hand 72.92\%. When area side length was decreased further to $l=5$cm, a clear failing point was reached: the left hand success rate dropped to 7.97\%, and the right hand to 20.94\%.

Besides the raw numbers, the collected data revealed characteristics related to the tool performance. Fig. \ref{fig:place_flow_success} shows the number of incorrectly evaluated area selections for each pointed area, $l=10$cm. The hatched areas are considered outliers, as they included out of bound gesturing and selections based on secondary snap criterion, Euclidean distance to area center points. 

Most of the successful selections are located diagonally over the selection areas. The dominant (right) hand performs better than the left hand. Mirrored patterns can be detected for both performance and the offsets that occur for incorrect area selections. While area size $l=5$cm can be considered a failing point, it provided similar data regarding the pointing offsets and performance.

\subsection{Integration analysis}
\begin{figure}[!b]
    \centering
    \includegraphics[width=9cm]{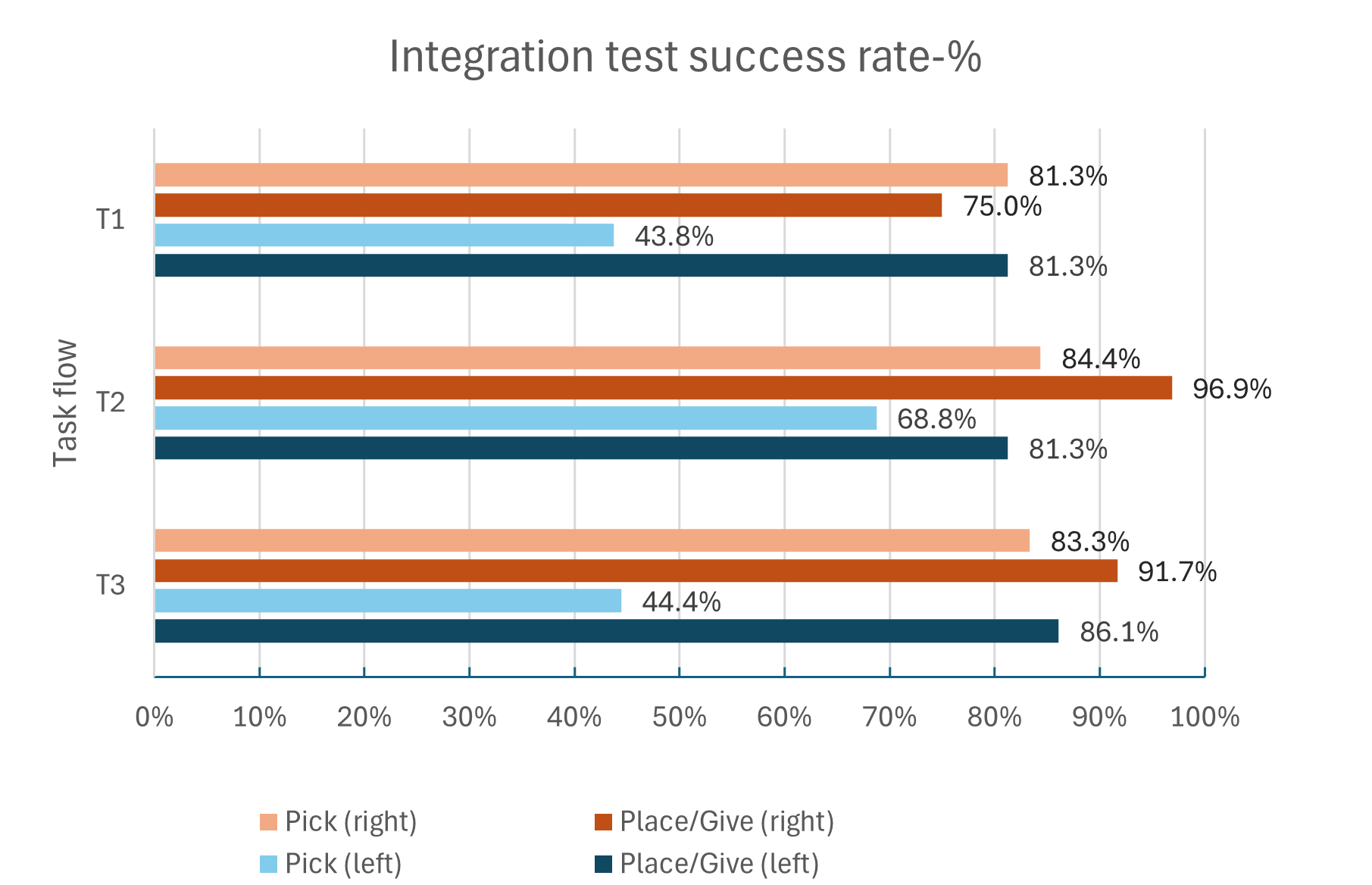}
    \caption{Integration test results.}\label{tab:integration_tests}
\end{figure}

\begin{figure*}[!t]
    \centering
    
    \includegraphics[width=7in]{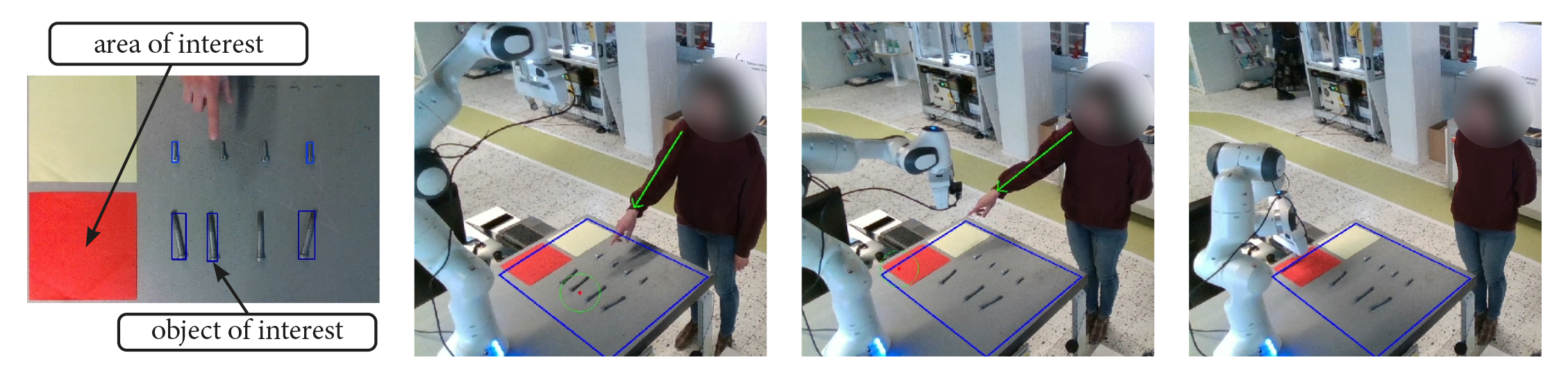}
    \caption{Integration task flow 1: Pick and Place sequence.}\label{fig:integration_place_sequence}
\end{figure*}

Integration results for task flows 1--3 are shown in Fig. \ref{tab:integration_tests}, and an example of pick-and-place sequence is illustrated in Fig.\ref{fig:integration_place_sequence}. The results suggest the dominant (right) hand performs adequately in the proposed tasks. The right hand pick success rate ranges between 81.3 to 84.4\%, while place and give success ranges between 75.0 to 96.9\%. The pick tasks are naturally more challenging for the system, as the placement of objects is close to accuracy limits of the gesturing tool. The left hand results are not as convincing: the pick success rate reaches 43.8--68.8\%, while place and give success rates are 81.3--86.1\%. For both hands, target group selection by speech improves the success rate. 

It is noteworthy, that object detection, control and grasping issues were present during the tests. Besides the occasional ghost detections, the object detection failed frequently classifying the objects correctly, demanding intermediate adjustments to the setup and causing pipeline failures. Grasping and control failures occurred a few times throughout the testing. Besides the mentioned pipeline failures, it is significant to acknowledge how the actions become gradually more successful as the targets decrease on the table, skewing the results. The integration tests do not provide perfectly reliable data for the gesturing-wise system performance, but lay the foundation for a proof-of-concept integration. Further testing is needed, to evaluate the system-wide performance more accurately.  

\subsection{Limitations} \label{subsec:limitations}

While pose estimation tools offer an easy alternative to implement and interpret pointing gestures in the scene, some deficiencies were notable during the integration tests. The original integration setup included an option to use both hands as a selection method, but
overlapping hands posed a challenge to the system. Considering the camera angle in Fig. \ref{fig:integration_place_sequence}, the issue arises when the workspace is pointed with the left hand and the right hand rests aside. The right wrist pose is incorrectly estimated due to overlapping left arm, causing right-handed ghost gestures on the plane.

The problem could be avoidable by changing the camera position, but locating the device in an optimal way is a tricky task, considering the occlusions created by the environment and the gesturing itself. As argued by works free of pose estimation \cite{shukla_probabilistic_2015, jirak_solving_2021}, skeletal estimation is a constraint when full-body or partial-body view is required to be visible during the interaction. Additionally, pose estimation is more sensitive to environmental factors compared to wearable sensor-based approaches. The implemented tool is not robust for multiple pose detections, which may occur naturally but also due to unfavourable workspace conditions, such as cluttered backgrounds and lightning.

The overlapping arms issue could be solved by developing more sophisticated selection strategies than the presented simple algorithms. Unlike works by \cite{droeschel_learning_2011, nickel_visual_2007}, the strategies do not consider well the temporal nature of gesturing sequences to recognize the active and dormant states of gestures. Recognizing a starting gesture would improve the integration and usability of the tool, and could enable both hands to be used simultaneously. On the other hand, improving the elbow-wrist configuration for the tool could reduce the needed field-of-view for the tool.

Area-wise performance differences could be addressed by investigating the offset data further for tool optimization. Testing could be improved by considering angular errors besides the Euclidean distance between the ground truth and targets. Calibration sequences such in \cite{jevtic_personalized_2019} could help the deployment of the tool and improve the performance. Studies on understanding the underlying problems in gesture interpretation \cite{herbort_spatial_2016} could provide further insight for optimizing the tool. As argued by \cite{herbort_spatial_2016}, different models for pointing gesture production and interpretation should be considered, to make human-technology interaction such as HRC more efficient. 

Finally, the gesturing tool was tested on a limited, table-like planar workspace. Many of the existing works \cite{hu_augmented_2022, lorentz_pointing_2023} have evaluate pointing gestures on floor level workspaces, which opens possibilities for wider range of applications. This remains to be tested for the current implementation. 

\section{Conclusion}\label{sec:conclusion}

Pointing gestures are a natural way to indicate areas of interest and guide robot's attention to desired targets in Human-Robot Collaboration. In this paper, the modality was explored using RGB-D camera, pose estimation and a simple geometric model to localize the pointed targets. The model reached an average accuracy of 3.0--3.3cm for the dominant hand, and 6.4--6.7cm for the non-dominant hand, when pointing ten targets on a table. The characteristics and shortcomings of the model were discussed through qualitative testing, and a proof-of-concept integration into a collaborative application was implemented. 

Future work includes studying the characteristics of the model further, to create an optimization and calibration sequence for leveling out area-wise performance differences. Solving occlusion issues algorithmically would increase the usability in Human-Robot Collaboration by giving operator freedom to use either of hands in actions. Elbow-wrist alternative and its optimization should be further explored, as it could reduce the field-of-view requirements for pose estimation. The gesturing tool basis can be harnessed beyond target selection for multimodal functionalities in collaborative scenarios. 

\section*{Acknowledgements}

This project has received funding from the European Union's Horizon Europe research and innovation programme under grant agreement no. 101059903 and 101135708.

\bibliographystyle{IEEEtran}
\bibliography{IEEEabrv,refs}

\end{document}